\documentclass{article}

\usepackage[preprint]{neurips}

\usepackage[utf8]{inputenc} % allow utf-8 input
\usepackage[T1]{fontenc}    % use 8-bit T1 fonts

\usepackage{url}            % simple URL typesetting
\usepackage{booktabs}       % professional-quality tables
\usepackage{amsfonts}       % blackboard math symbols
\usepackage{nicefrac}       % compact symbols for 1/2, etc.
\usepackage{microtype}      % microtypography
\usepackage{xcolor}         % colors
\usepackage{graphicx}
\usepackage{graphicx,wrapfig}
\usepackage{amsmath}
\usepackage{amssymb}
\usepackage{mathtools}
\usepackage{amsthm}
\usepackage{xcolor}
\usepackage[numbers]{natbib}
\usepackage{multirow}
\usepackage{hyperref}       % hyperlinks

\theoremstyle{plain}
\newtheorem{theorem}{Theorem}[section]
\theoremstyle{definition}
\newtheorem{definition}[theorem]{Definition}

\title{Leader-Follower Neural Networks with Local Error Signals Inspired by Complex Collectives}

\author{%
  Chenzhong Yin \\
  University of Southern California\\
  Los Angeles, CA 90089 \\
  \And
  Mingxi Cheng \\
  University of Southern California\\
  Los Angeles, CA 90089 \\
  \AND
  Xioingye Xiao \\
  University of Southern California\\
  Los Angeles, CA 90089 \\
  \And
  Xinghe Chen \\
  University of Southern California\\
  Los Angeles, CA 90089 \\
  \And
  Shahin Nazarian\\
  University of Southern California\\
  Los Angeles, CA 90089 \\
  \And
  Andrei Irimia\\
  University of Southern California\\
  Los Angeles, CA 90089 \\
  \And
  Paul Bogdan\\
  University of Southern California\\
  Los Angeles, CA 90089 \\
}

\begin{document}

\maketitle

\begin{abstract}
The collective behavior of a network with heterogeneous, resource-limited information processing units (e.g., group of fish, flock of birds, or network of neurons) demonstrates high self-organization and complexity. These emergent properties arise from simple interaction rules where certain individuals can exhibit leadership-like behavior and influence the collective activity of the group. 
Motivated by the intricacy of these collectives, we propose a neural network (NN) architecture inspired by the rules observed in nature's collective ensembles. This NN structure contains \textit{workers} that encompass one or more information processing units (e.g., neurons, filters, layers, or blocks of layers). Workers are either leaders or followers, and we train a leader-follower neural network (LFNN) by leveraging local error signals and optionally incorporating backpropagation (BP) and global loss. We investigate worker behavior and evaluate LFNNs through extensive experimentation. 
Our LFNNs trained with local error signals achieve significantly lower error rates than previous BP-free algorithms on MNIST and CIFAR-10 and even surpass BP-enabled baselines. In the case of ImageNet, our LFNN-$\ell$ demonstrates superior scalability and outperforms previous BP-free algorithms by a significant margin.
% On MNIST and CIFAR-10, our LFNNs trained with local error signals achieve remarkably lower error rates when compared to previous algorithms that do not rely on BP, and even surpasses BP-enabled baselines. Our LFNN-$\ell$ also scales well and significantly outperforms previous BP-free algorithms on ImageNet.
\end{abstract}

\section{Introduction}
Artificial neural networks (ANNs) typically employ global error signals for learning \cite{rumelhart1985learning}. While ANNs draw inspiration from biological neural networks (BNNs), they are not exact replicas of their biological counterparts.
ANNs consist of artificial neurons organized in a structured layered architecture~\cite{thomas2008connectionist}. Learning in such architectures commonly involves gradient descent algorithms~\cite{bottou1991stochastic} combined with backpropagation (BP) \cite{rojas1996backpropagation}. Conversely, BNNs exhibit more intricate self-organizing connections, relying on specific local connectivity \cite{markram2011history} to enable emergent learning and generalization capabilities even with limited and noisy input data. 
Simplistically, we can conceptualize a group of neurons as a collection of \textit{workers} wherein each worker receives partial information and generates an output, transmitting it to others so as to achieve a specific collective objective. 
This behavior can be observed in various biological systems, such as decision-making among a group of individuals \cite{moscovici1969group}, flocking behavior in birds to avoid predators and maintain flock health \cite{o1999alternating}, or collective behavior in cells fighting infections or sustaining biological functions \cite{friedl2004collective}.

The study of collective behavior in networks of heterogeneous agents, ranging from neurons and cells to animals, has been a subject of research for several decades. In physical systems, interactions among numerous particles give rise to emergent and collective phenomena, such as stable magnetic orientations \cite{hopfield1982neural}. A system of highly interconnected McCulloch-Pitts neurons \cite{mcculloch1943logical} has collective computational properties~\cite{hopfield1982neural}. Networks of neurons with graded response (or sigmoid input-output relation) exhibit collective computational properties similar to those of networks with two-state neurons~\cite{hopfield1984neurons}.
Recent studies focus on exploring collective behaviors in biological networks. This includes the examination of large sensory neuronal networks~\cite{tkavcik2014searching}, the analysis of large-scale small-world neuronal networks~\cite{qu2017collective}, the investigation of heterogeneous NNs \cite{luccioli2010irregular}, and the study of hippocampal networks \cite{meshulam2017collective}. These studies aim to uncover the collective dynamics and computational abilities exhibited by such biological networks.
% More modern studys have also been proposed to search for collective behaviors in biological networks, such as large networks of sensory neurons \cite{tkavcik2014searching}, large-scale small-world neuronal networks and \cite{qu2017collective}, heterogeneous neural networks \cite{luccioli2010irregular}, hippocampal networks with place and non-place neurons \cite{meshulam2017collective}.

\begin{figure}[!t]
  \centering
  \vspace{-0.3cm}
  \includegraphics[width=\columnwidth]{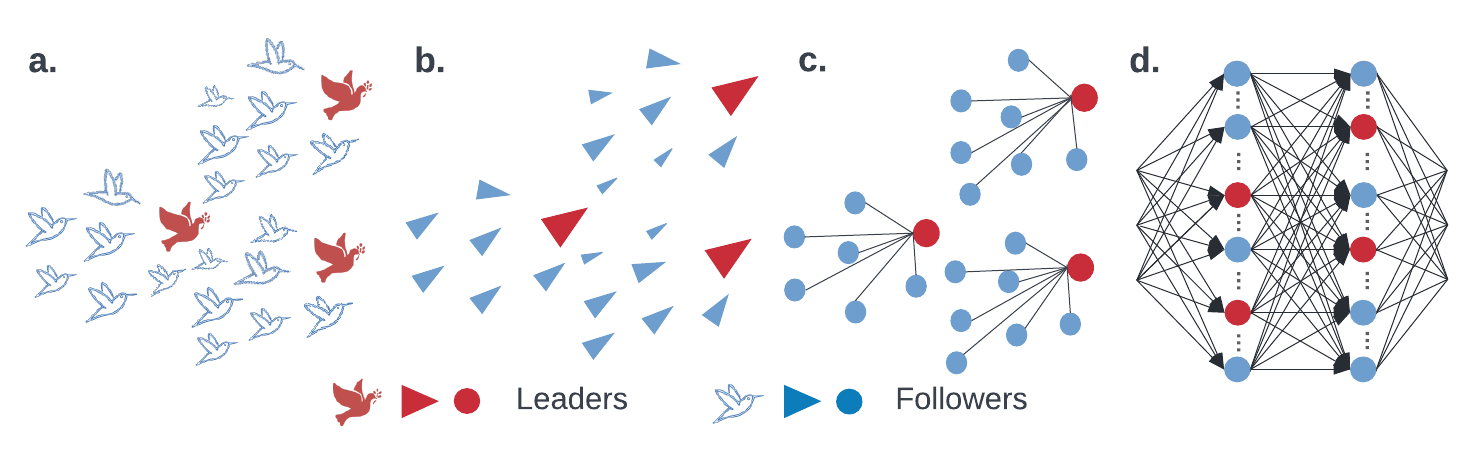}
  \vspace{-0.7cm}
  \caption{ \textbf{a-b.} A flock of birds where leaders are informed and lead the flock.  \textbf{c.} An abstracted network from the flock.  \textbf{d.} A leader-follower neural network architecture.}
  \vspace{-0.5cm}
  \label{fig:flock}
\end{figure}

In biological networks such as the human brain, synaptic weight updates can occur through local learning, independent of the activities of neurons in other brain regions~\cite{caporale2008spike, yin2023anatomically}. Partly for this reason, local learning has been identified as effective means to reduce memory usage during training and to facilitate parallelism in deep learning architectures, thereby enabling faster training~\cite{narayanan2019pipedream, xiong2020loco}. 
%Xiong et al. introduced a local learning approach that combines two blocks into a single local block, simultaneously shared by two units, to bridge the local loss blocks~\cite{xiong2020loco}.

% Collective motion and goal-oriented swarming represent extensively studied phenomena in complex biological systems. To better understand and characterize these behaviors at a microscopic level, various mathematical models have been developed. One prominent example is the Vicsek model \cite{vicsek1995novel}, which offers a minimalistic set of equations to describe collective motion. This model exhibits a phase transition from disordered motion to large-scale ordered motion, depending on factors such as the strength of interactions among individuals and the magnitude of noisy perturbations \cite{gregoire2004onset}. The Vicsek model serves as a valuable tool for investigating and analyzing the emergence of coordinated motion in swarms.
% Among the numerous phenomena exhibited by complex biological systems, collective motion and goal-oriented swarming are the most studied. For this purpose, several mathematical models have been developed to describe such behaviors from a microscopic description. For instance, the Vicsek model \cite{vicsek1995novel} provides a minimalistic set of equations to describe collective motion and exhibits a phase transition from disordered motion to large-scale ordered motion as a function of interaction strength and magnitude of noisy perturbations \cite{gregoire2004onset}.
% \textcolor{black}{please state the main findings here}. 

Acknowledging the extensive research on collective behavior and local learning in biological networks and the existing computational disparities between ANNs and BNNs, we draw inspiration from complex collective systems by proposing a NN architecture with more complex capabilities observed in biological counterparts. 
We propose to divide a NN into layers of elementary \textit{leader} workers and \textit{follower} workers and follow the characteristics of collective motion to select the \textit{leadership}.
As in a flock of birds shown in Figure\ref{fig:flock}, leaders are informed by and control the motion of the whole flock.
In our \textit{leader-follower neural network} (LFNN) architecture, the leaders and followers differ in their access to information and learn through distinct error signals. Hence, LFNNs offer a biologically-plausible alternative to BP and facilitate training using local error signals.
% are re-selected in each epoch and they are informed by directly receiving error signals for weight update. 
% In contrast, the \textit{follower} workers do not get the same loss signals; instead, they only try to follow leaders' steps and move closer to leaders. 
% In other words, neurons in our network learn with different local signals and only a small portion of neurons (representing the leadership) are informed with prediction loss.

We evaluated our LFNN and its BP-free version trained with local loss (LFNN-$\ell$) on MNIST, CIFAR-10, and ImageNet datasets. Our LFNN and LFNN-$\ell$ outperformed other biologically plausible BP-free algorithms and achieved comparable results to BP-enabled baselines. Notably, our algorithm demonstrated superior performance on ImageNet compared to all other BP-free baselines. This study, which introduces complex collectives to deep learning, provides valuable insights into biologically plausible NN research and opens up avenues for future work.
% LFNNs achieve results similar to those of conventional NNs trained with BP and global loss. 
% For BP-free LFNN trained with local loss (LFNN-$\ell$, where $\ell$ stands for local), we provide extensive comparison with the state-of-the-art using three benchmark datasets and demonstrate that our LFNN-$\ell$ outperforms all baselines. %\textcolor{red}{But what does this architecture achieve that's BETTER than the state of the art? Reviewers could say that unless you can do something better, a simpler and more parsimonious model is preferable. }

\textbf{Related work.}
Efforts have been made to bridge the gaps in computational efficiency that continue to exist between ANNs and BNNs \cite{bartunov2018assessing}. One popular approach is the replacement of global loss with local error signals \cite{mostafa2018deep}. Researchers have proposed to remove BP to address backward locking problems \cite{jaderberg2017decoupled}, mimic the local connection properties of neuronal networks \cite{pyeon2020sedona} and incorporate local plasticity rules to enhance ANN's biological plausibility \cite{illing2021local}.
A research topic closely related to our work is supervised deep learning with local loss. 
It has been noticed that training NNs with BP is biologically implausible because BNNs in the human brain do not transmit error signals at a global scale \cite{crick1989recent,marblestone2016toward,lillicrap2020backpropagation}. 
% Alternatives have been proposed to replace BP and solve the \emph{locking problems} accordingly. 
Several studies have proposed training NNs with local error signals, such as layer-wise learning \cite{mostafa2018deep,nokland2019training}, block-wise learning \cite{pyeon2020sedona,lowe2019putting}, gated linear network family \cite{veness2019gated}, etc. Mostafa et al. generate local error signals in each NN layer using fixed, random auxiliary classifiers \cite{mostafa2018deep}, where a hidden layer is trained using local errors generated by a random fixed classifier. This is similar to an approach called feedback alignment training, where random fixed weights are used to back-propagate the error layer by layer \cite{lillicrap2016random}. 
In \cite{lowe2019putting}, the authors split a NN into a stack of gradient-isolated modules, and each module is trained to maximally preserve the information of its inputs. 
A more recent work by Ren et al.~\cite{ren2022scaling} proposed a local greedy forward gradient algorithm by enabling the use of forward gradient learning in supervised deep learning tasks. Their biologically plausible BP-free algorithm outperforms the forward gradient and feedback alignment family of algorithms significantly. 
Our LFNN-$\ell$ shares some similarities with the above work in the sense that the LFNN-$\ell$ is trained with loss signals generated locally without BP. In contradistinction to the state-of-the-art, we do not require extra memory blocks to generate an error signal. Hence, the number of trainable parameters can be kept identical to that of NNs without an LF hierarchy.

\section{LFNNs Inspired by Complex Collectives} \label{sec:methods}
Collective motion refers to ordered movement in systems consisting of self-propelled particles, such as flocking \cite{o1999alternating} or swarming behavior \cite{bouffanais2016design}. The main feature of such behavior is that an individual particle is dominated by the influence of others and thus behaves entirely differently from how it might behave on its own~\cite{vicsek2012collective}. 
A classic collective motion model, the Vicsek model~\cite{vicsek1995novel}, describes the trajectory of an individual using its velocity and location, and uses stochastic differential/ difference equations to update this agent's location and velocity as a function of its interaction strength with its neighbors.
% and the magnitude of noise perturbations.
%at each time, (i) an individual particle aligns with its neighbors to update its velocity, and (ii) the particle then updates its location by following the new velocity. 
Inspired by collective motion seen in nature, we explore whether these minimal mathematical relations can be exploited in deep learning. 

\begin{figure}[!t]
  \centering
  \vspace{-0.3cm}
  \includegraphics[width=0.95\columnwidth]{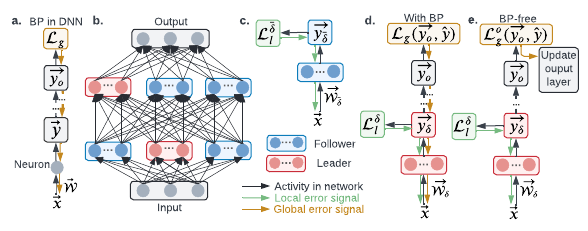}
  \vspace{-0.5cm}
  \caption{\textbf{Weight updates of LFNN.} \textbf{a.} BP in classic deep neural network (DNN) training. Global prediction loss is back-propagated through layers.  \textbf{b.} An LF hierarchy in a DNN. Within a layer, neurons are grouped as (leader and follower) workers.  \textbf{c.} Weight update of follower workers.  \textbf{d.} Weight update of leader workers with BP. \textbf{e.} BP-free weight update of leader workers.}
  \vspace{-0.5cm}
  \label{fig:weightupdate}
\end{figure}

\textbf{LF hierarchy in fully connected layers.}
% We consider a NN with multiple fully-connected (FC) layers, and each layer contains several neurons. 
In a fully-connected (FC) layer containing multiple neurons, we define \textit{workers} as structures containing one or more neurons grouped together. In contradistinction to classic NNs where the neuron is the basic computational unit, LFNN workers serve as basic units. By adapting the Vicsek model terms to deep learning, a worker's behavior is dominated by that of neighbors in the same layer. In addition, we consider \textit{leadership} relations inside the group. According to collective motion, ``leadership'' involves ``the initiation of new directions of locomotion by one or more individuals, which are then readily followed by other group members'' \cite{krause2000leadership}. Thus, in FC layers, one or more workers are selected as leaders, and the rest are ``followers'' as shown in Figure \ref{fig:weightupdate}b. 

\textbf{LF hierarchy extended in convolutional layers. }Given a convolutional layer with multiple filters (or kernels), workers can be defined as one or more filters grouped together to form \textit{filter-wise workers}. For a more coarsely-grained formulation, given a NN with multiple convolutional layers, a set of convolutional layers can be grouped naturally as a block (such as in VGG~\cite{simonyan2014very}, ResNet~\cite{he2016deep}, Inception~\cite{szegedy2015going} architectures). Our definition of the worker can be easily adapted to encompass \textit{block-wise workers} to reflect this architecture where a block of convolutional layers work together as a single, block-wise worker. Similarly, if a block contains one layer, it becomes a \textit{layer-wise worker}. 

More formally, we consider a NN with $\mathcal{M}$ hidden layers, and a hidden layer contains $\mathcal{N}$ workers. A worker can contain one or more individual working components, which can be neurons, filters in convolutional layers, or blocks of NN layers, and each individual working component is parametrized by a set of trainable parameters $\mathcal{W}$. During training, at each time step $t$, leader workers $\mathcal{N}_{\delta}$ are dynamically selected and the remaining workers are labeled as followers (denoted as $\mathcal{N}_{\bar \delta}$) at time step $t$. Following the same notation, leader and follower workers are parameterized by matrices $\vec{\mathcal{W}}_{\delta}$ and $\vec{\mathcal{W}}_{\bar \delta}$, respectively. The output of leader and follower workers in a hidden layer reads $f(\vec{x},[\vec{\mathcal{W}}_{\delta},\vec{\mathcal{W}}_{\bar\delta}])$, where $\vec{x}$ is the input to the current hidden layer and $f(\cdot)$ is a mapping function.

\textbf{Error signals in LFNN. }In human groups, one key difference between leaders and followers is that leaders are \textit{informed} individuals that can guide the whole group, while followers are uninformed and their instructions differ from treatment to treatment \cite{faria2010leadership}. 
Adapting this concept to deep learning, LFNN leaders are informed that they receive error signals generated from the global or local prediction loss functions, whereas followers do not have this information. 
Specifically, assume that we train an LFNN with BP and a global prediction loss function $\mathcal{L}_g$. Only leaders $\mathcal{N}_{\delta}$ and output neurons receive information on gradients as error signals to update weights. This is similar to classic NN training, so we denote these pieces of information as \textit{global error signals}. In addition, a local prediction error $\mathcal{L}_l^{\delta}$ is optionally provided to leaders to encourage them to make meaningful predictions independently. 

By contrast to leaders, followers $\mathcal{N}_{\bar\delta}$ do not receive error signals generated in BP. Instead, they align with their neighboring leaders. 
Inspired by collective biological systems, we propose an ``alignment'' algorithm for followers and demonstrate its application in an FC layer as follows: Consider an FC layer where the input to a worker is represented by $\vec{x}$, and the worker is parameterized by $\vec{\mathcal{W}}$ (i.e., the parameters of all neurons in this worker). The output of a worker is given by $\vec{y} = f(\vec{\mathcal{W}}\cdot\vec{x})$. 
In this context, we denote the outputs of a leader and a follower as $\vec{y}_\delta$ and $\vec{y}_{\bar\delta}$, respectively. To bring the followers closer to the leaders, a local error signal is applied to the followers, denoted as $\mathcal{L}_l^{\bar\delta} = \mathcal{D}(\vec{y}_\delta,\vec{y}_{\bar\delta})$, where $\mathcal{D}(a,b)$\footnote{In our experimentation, we utilize mean squared error loss.} measures the distance between $a$ and $b$.
In summary, the loss function of our LFNN is defined as follows:
\begin{equation}
\label{eq:bp}
    \mathcal{L} = \mathcal{L}_g +\lambda_1 \mathcal{L}_l^{\delta} + \lambda_2 \mathcal{L}_l^{\bar\delta},
\end{equation}
where the first term of the loss function applies to the output neurons and leader workers. The second and third terms apply to the leader and follower workers, as illustrated in Figure \ref{fig:weightupdate}c and d. The hyper-parameters $\lambda_1$ and $\lambda_2$ are used to balance the contributions of the global and local loss components. It is important to note that the local loss $\mathcal{L}_l^{\delta}$ and $\mathcal{L}_l^{\bar\delta}$ are specific to each layer, filter, or block and do not propagate gradients through all hidden layers.

\textbf{BP-free version (LFNN-$\ell$). }
To address the limitations of BP such as backward locking, we propose a BP-free version of LFNN. The approach is as follows: In Eq. \ref{eq:bp}, it can be observed that the weight updates for followers are already local and do not propagate through layers. Based on this observation, we modify LFNN to train in a BP-free manner by removing the BP for global prediction loss. Instead, we calculate leader-specific local prediction loss ($\mathcal{L}_l^{\delta}$) for all leaders.
This modification means that the global prediction loss calculated at the output layer, denoted as $\mathcal{L}_g^o$ (where $o$ stands for output), is only used to update the weights of the output layer. In other words, this prediction loss serves as a local loss for the weight update of the output layer only.
The total loss function of the BP-free LFNN-$\ell$ is given as follows:
\begin{equation}
\label{eq:bpfree}
    \mathcal{L} = \mathcal{L}_g^o + \mathcal{L}_l^{\delta}+ \lambda \mathcal{L}_l^{\bar\delta}.
\end{equation}
By eliminating the backpropagation of the global prediction loss to hidden layers, the weight update of leader workers in LFNN is solely driven by the local prediction loss, as depicted in Figure \ref{fig:weightupdate}e. It's important to note that the weight update of follower workers remains unchanged, regardless of whether backpropagation is employed or not, as shown in Figure \ref{fig:weightupdate}c.

\textit{Dynamic leadership selection. }In our LF hierarchy, the selection of leadership is dynamic and occurs in each training epoch based on the local prediction loss. In a layer with $\mathcal{N}$ workers, each worker can contain one or more neurons, enabling it to handle binary or multi-class classification or regression problems on a case-by-case basis. This unique characteristic allows a worker, even if it is located in hidden layers, to make predictions $\vec{y}$. This represents a significant design distinction between our LFNN and a traditional neural network. 
Consequently, all workers in a hidden layer receive their respective prediction error signal, denoted as $\mathcal{L}_l^\delta(\vec{y}, \hat{y})$. Here, $\mathcal{L}_l(\cdot,\cdot)$ represents the prediction error function, the superscript $\delta$ indicates that it is calculated over the leaders, $\hat{y}$ denotes the true label, and the top $\delta$ ($0\leq\delta\leq100\%$) workers with the lowest prediction error are selected as leaders. 
\begin{definition}[Leadership] \label{def:leader}
    Within a set of $\mathcal{N}$ workers, each worker generates a prediction error denoted as $\mathcal{L}_l(\vec{y}, \hat{y})$. From this set, we select $\delta$ \textit{leaders} based on their lowest prediction errors. The prediction loss for these leaders is represented as $\mathcal{L}_l^\delta(\vec{y}, \hat{y})$. The remaining workers are referred to as \textit{followers}, and their prediction loss is denoted as $\mathcal{L}_l^{\bar\delta}(\vec{y}, \hat{y})$.
\end{definition}

\textit{Implementation details.} To enable workers in hidden layers to generate valid predictions, we apply the same activation function used in the output layer to each worker. For instance, in the case of a neural network designed for $K$-class classification, we typically include $K$ output neurons in the output layer and apply the softmax function. In our LFNN, each worker is composed of $K$ neurons, and the softmax function is applied accordingly.
% Different than leaders, which are informed and follow prediction loss signals to update weights, followers are designed to align with \textit{neighboring} leaders. 
% In fully connected layers, all workers in the current layer are connected with all workers from the previous layer, hence, workers in the same layer are all \textit{neighbors}. 
In order to align the followers with the leaders, we adopt a simplified approach by selecting the best-performing leader as the reference for computing $\mathcal{L}_l^{\bar\delta}$. While other strategies such as random selection from the $\delta$ leaders were also tested, they did not yield satisfactory performance. Therefore, for the sake of simplicity and better performance, we choose the best-performing leader as the reference for the followers' loss computation.

\textit{Practical benefits and overheads.} In contrast to conventional neural networks trained with BP and a global loss, our LFNN-$\ell$ computes worker-wise loss and gradients locally. This approach effectively eliminates backward locking issues, albeit with a slight overhead in local loss calculation.
One significant advantage of the BP-free version is that local error signals can be computed in parallel, enabling potential speed-up in the weight update process through parallel implementation.

\section{Experiments}
% In this section, we conduct experiments to evaluate the performance of our proposed LFNN and LFNN-$\ell$. 
In Section \ref{sec: dnn}, we focus on studying the leadership size, conducting an ablation study of loss terms in Eq. \ref{eq:bp}, and analyzing the worker's activity. To facilitate demonstration and visualization, we utilize DNNs in this subsection. 
In Section \ref{sec:LFNN-l}, we present our main experimental results, where we evaluate LFNNs and LFNN-$\ell$s using CNNs on three datasets (i.e., MNIST, CIFAR-10, and ImageNet) and compare with a set of baseline algorithms.

\begin{figure}[!t]
  \centering
  \vspace{-0.3cm}
  \includegraphics[width=\columnwidth]{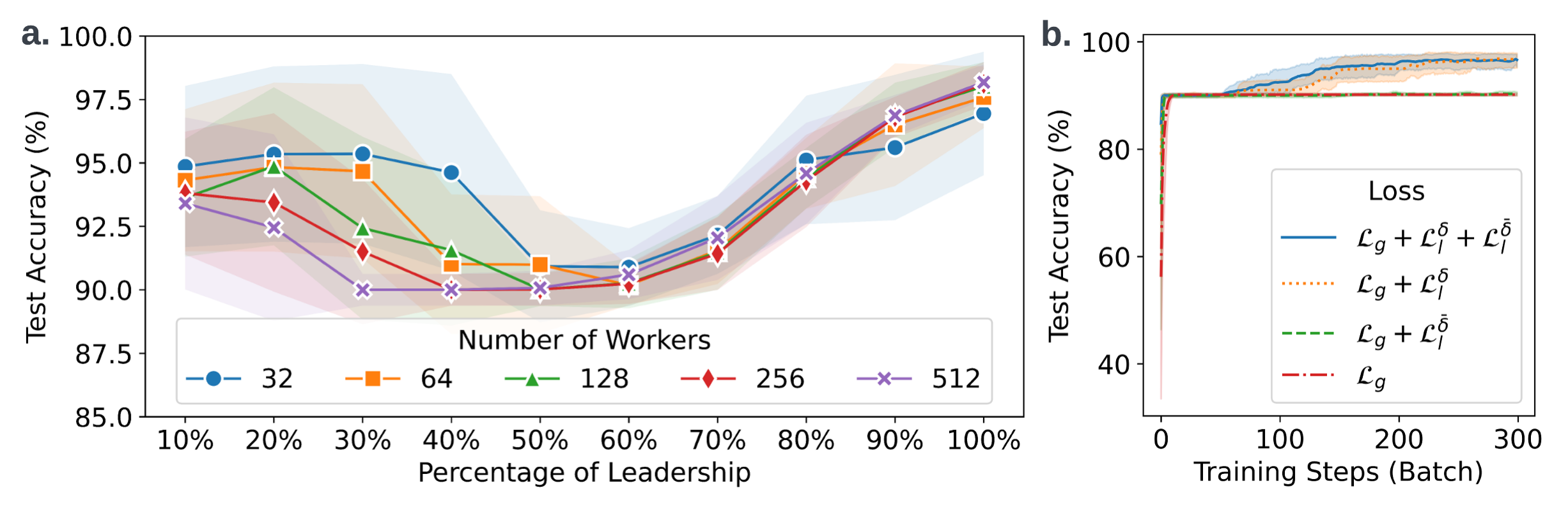}
  \vspace{-0.5cm}
  \caption{\textbf{a.} Network performance results when varying leadership size from 10\% to 100\%. \textbf{b.} Ablation study results  from four different loss functions.}
  \vspace{-0.7cm}
  \label{fig:pmnist}
\end{figure}

\subsection{Leader-Follower Neural Networks (LFNNs)} \label{sec: dnn}
\textbf{Experimental setup. } 
To assess the performance of LFNN for online classification, we conduct experiments on the pixel-permuted MNIST dataset \cite{lecun1998mnist}. Following the approach in \cite{veness2019gated}, we construct a one-vs-all classifier using a simple neural network architecture consisting of one hidden FC layer. In our experiments, we vary the network architecture to examine the relationship between network performance and leadership size. We consider network configurations with 32, 64, 128, 256, and 512 workers, where each worker corresponds to a single neuron. We systematically vary the percentage of workers assigned as leaders from 10\% to 100\%.
For each network configuration, we utilize the sigmoid activation function for each worker and train the model using the Adam optimizer with a learning rate of 5e-3. The objective is to investigate how different leadership sizes impact the classification performance in the online setting.
% \textcolor{black}{All experiments in this manuscript are conducted with Google Colab Pro (Nvidia P100/T40 GPU and 32GB RAM).}
In our experiments, we employ the binary cross-entropy loss for both the global prediction loss ($\mathcal{L}_g$) and the local prediction loss for leaders ($\mathcal{L}_l^\delta$). For the local error signal of followers ($\mathcal{L}_l^{\bar\delta}$), we use the mean squared error loss. The hyperparameters $\lambda_1$ and $\lambda_2$ are both set to 1 in this section to balance the global and local loss terms.
In the ablation study of loss terms and the worker activity study, we focus on a 32-worker LFNN with 30\% leadership. 

\textbf{Leadership size and performance.} 
In a study on the collective motion of inanimate objects, such as radio-controlled boats, it was observed that to effectively control the direction of the entire group, only a small percentage (5\%-10\%) of the boats needed to act as leaders \cite{tarcai2011patterns}. This finding aligns with similar studies conducted on living collectives, such as fish schools and bird flocks, where a small subset of leaders were found to have a substantial impact on the behavior of the larger group.
In our experiment, we investigate the relationship between network performance and the size of the leadership group. The results shown in Figure \ref{fig:pmnist}a indicate that our LFNN achieves high performance on the permuted MNIST classification task after just one pass of training data. When using a higher percentage of leadership, such as 90\% or 100\%, the LFNN achieves comparable performance to a DNN trained with BP. Even with a lower percentage of leadership, ranging from 10\% to 30\%, the LFNN still achieves decent performance on this task. It is worth noting that for more challenging datasets like ImageNet, higher percentages of leadership are preferred. These findings highlight both the similarities and differences between natural collectives and LFNNs in the field of deep learning.  

% \textcolor{blue}{mingxi: can remove if no space. }Surprisingly, when trained with BP, the relationship between the performance of the LFNN and leadership size is not positively correlated. It is not the case that the higher the leadership percentage, the higher the accuracy. In fact, we observe that in our experiment, the network performs the worst when the leader and follower are about half and half of the population. We hypothesize that since leadership in the one-pass training process is dynamically changing, and when workers follow constantly changing leaders, the network is not dominated by either followers or leaders, and hence, it does not move toward a direction and could be trapped in saddle points.

\begin{figure}[!t]
  \centering
  \vspace{-0.3cm}
  \includegraphics[width=\columnwidth]{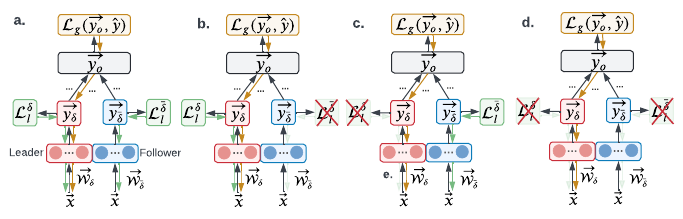}
  \vspace{-0.7cm}
  \caption{\textbf{Loss variation demonstration.} \textbf{a.} Global prediction loss and both local losses, $\mathcal{L}_1$. \textbf{b.} Without local follower loss, $\mathcal{L}_2$. \textbf{c.} Without local leader loss, $\mathcal{L}_3$. \textbf{d.} Global prediction loss alone, $\mathcal{L}_4$.}
  \vspace{-0.5cm}
  \label{fig:ablation study of loss}
\end{figure}

\textbf{Ablation study of loss terms. }In our investigation of LFNN training using Eq. \ref{eq:bp}, we aim to evaluate the effectiveness of the local loss terms and examine the following aspects in this section: (a) whether global loss alone with BP is adequate for training LFNNs, and (b) how the inclusion of local losses contributes to training and network performance in terms of accuracy.
To address these questions, we consider four variations of the loss function, as depicted in Figure~\ref{fig:ablation study of loss}: 
(i) $\mathcal{L}_1 = \mathcal{L}_g + \mathcal{L}_l^{\delta} + \mathcal{L}_l^{\bar\delta}$: This variant includes the global loss as well as all local losses.
(ii) $\mathcal{L}_2 = \mathcal{L}_g + \mathcal{L}_l^{\delta}$: Here, the global loss is combined with the local leader loss.
(iii) $\mathcal{L}_3 = \mathcal{L}_g + \mathcal{L}_l^{\bar\delta}$: This variant utilizes the global loss along with the local follower loss.
(iv) $\mathcal{L}_4 = \mathcal{L}_g$: In this case, only the global loss is employed.
% By examining these different loss configurations, we aim to gain insights into the necessity and impact of local losses on the training process and overall network accuracy.

After training LFNNs with the four different loss functions mentioned earlier, we observe the one-pass results in Figure \ref{fig:pmnist}b. It is evident that using only the global prediction loss ($\mathcal{L}_4$) with backpropagation leads to the worst performance. The network's accuracy does not improve significantly when adding the local follower loss ($\mathcal{L}_3$) because the leader workers, which the followers rely on for weight updates, do not perform well. As a result, the overall network accuracy remains low.
However, when we incorporate the local leader loss ($\mathcal{L}_2$), we notice a significant improvement in the network's performance after 100 training steps. The local leader loss plays a crucial role in this improvement. Despite updating only 30\% of the workers at each step, it is sufficient to guide the entire network towards effective learning.
Moreover, when we further include the local follower loss ($\mathcal{L}_1$) to update the weights of followers based on strong leaders, the overall network performance improves even further. As a result, the network achieves high accuracy with just one pass of training data.
These results highlight the importance of incorporating both local leader and local follower losses in LFNN training. The presence of strong leaders positively influences the performance of followers, leading to improved network accuracy.

\textbf{Worker activity in an LFNN. }Collective motion in a group of particles is easily identifiable through visualization. Since our LFNN's weight update rules are inspired by a collective motion model, we visualize the worker activities and explore the existence of collective motion patterns in the network during training. Following our weight update rule, we select 30\% of the leaders from the 32 workers in each training step and update their weight dynamics based on global and local prediction loss. Consequently, the leader workers receive individual error signals and update their activity accordingly. Conversely, the remaining 70\% of workers act as followers and update their weight dynamics by mimicking the best-performing leader through local error signals. In essence, all followers align themselves with a single leader, resulting in similar and patterned activity in each training step. 
% Moreover, workers take larger steps when there is a significant difference between the output of leaders and workers.

\begin{figure}[!t]
  \centering
  \vspace{-0.3cm}
  \includegraphics[width=\columnwidth]{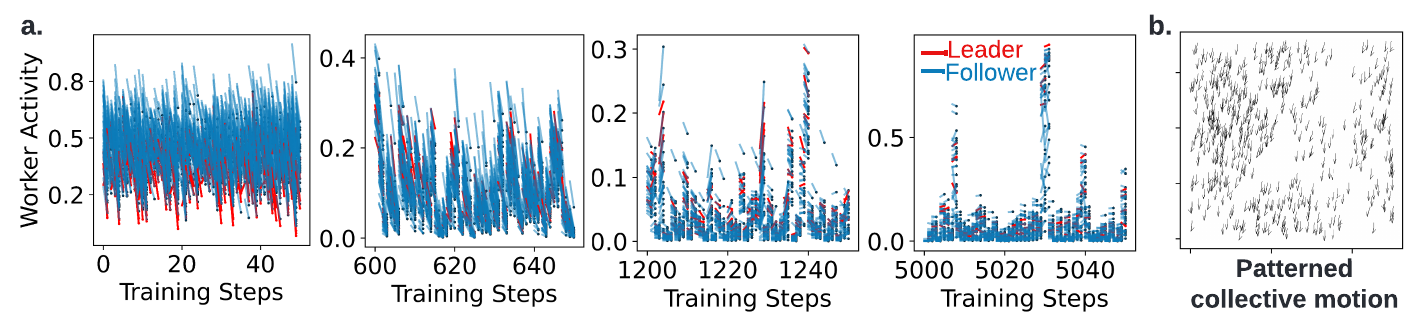}
  \vspace{-0.7cm}
  \caption{\textcolor{black}{\textbf{a.} Worker activity visualization in an LFNN. At each time step, the followers (blue lines) align themselves with leaders (red lines). \textbf{b.} Patterned collective motion produced by the classic Vicsek model \cite{vicsek1995novel}.}}
  \vspace{-0.3cm}
  \label{fig:collective motion}
\end{figure}

To visualize the activities of all workers, we utilize the neuron output $\vec{y}$ before and after the weight update at each time step, and the difference between them represents the worker activity. The results in Figure \ref{fig:collective motion}a demonstrate that in each time step, the follower workers (represented by blue lines) move in unison to align themselves with the leaders. During the initial training period (steps 0 to 1000), both leaders and followers exhibit significant movement and rapid learning, resulting in relatively larger step sizes. As the learning process stabilizes and approaches saturation, the workers' movement becomes less pronounced as the weights undergo less drastic changes in the well-learned network. Overall, we observe a patterned movement in worker activity in LFNNs, akin to the collective motion observed in the classic Vicsek model \cite{vicsek1995novel}.

\begin{figure}[!t]
  \centering
  % \vspace{-0.1cm}
  \includegraphics[width=\columnwidth]{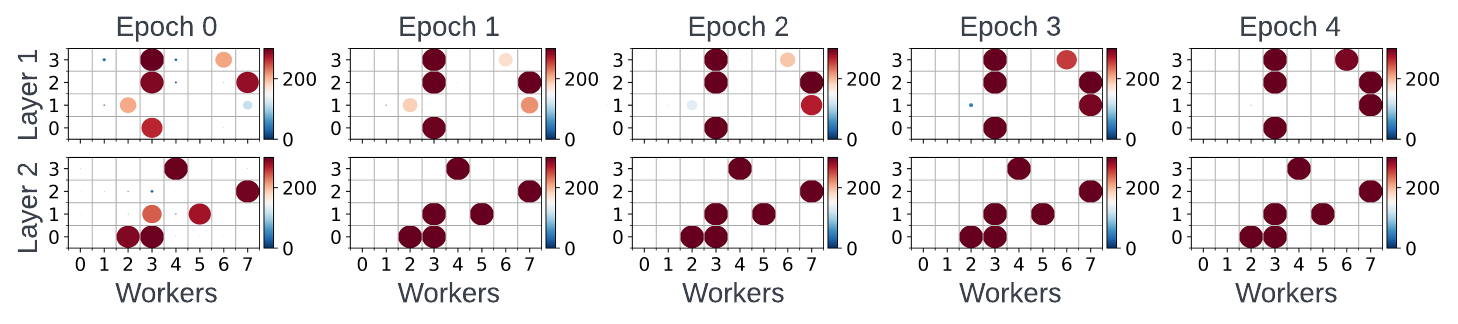}
  \vspace{-0.7cm}
  \caption{Leadership in workers during training. The color and size of thedots represent the times a worker is selected as leader. A worker can be selected as a leader up to 300 times in each epoch. }
  \vspace{-0.6cm}
  \label{fig:leadership}
\end{figure}

 \textbf{Leadership development.}
% Leaders in a leader-follower network are those workers with the lowest local loss values, and 
% they are selected dynamically in each training epoch. More specifically, 
In order to investigate how leadership is developed during training, we conduct a study using batch training, where leaders are re-selected in each batch. To provide a clearer demonstration, we focus solely on local losses in this study, thereby eliminating the effect of the global error signal and BP. We utilize an LFNN-$\ell$ with two hidden FC layers, each containing 32 workers. The leadership rate is fixed at 20\%, resulting in approximately 6 leaders being selected in each layer at every training step. The neural network is trained for 300 steps in each epoch, and the visualization of the leadership dynamics during the first 5 epochs is presented in Figure \ref{fig:leadership}.
In Figure \ref{fig:leadership}, the visualization depicts the development of leadership during training. Each dot's color and size indicate the number of times a worker is selected as a leader. In the initial epoch (Epoch 0), we observe that several workers in each layer have already emerged as leaders, being selected most of the time. As training progresses, exactly six workers in each layer are consistently developed as leaders, while the remaining workers are no longer selected. By the fifth epoch, the leadership structure becomes nearly fixed, remaining relatively unchanged throughout the training process.

From the results obtained, leadership in LFNN-$\ell$ is developed in the early stages of training and becomes fixed thereafter. The performance of the entire network relies on these leaders. 
% Surprisingly, even with a small percentage of leadership, the neural network achieves high accuracy, suggesting that the network capacity is more than sufficient for the given classification problem. 
% Although this aspect is not the primary focus of the current work, it presents an interesting avenue for future exploration, particularly in understanding the relationship between network size and dataset complexity.
Although this aspect is not the primary focus of the current work, one promising future direction involves the development of an intelligent dynamic leader selection algorithm.
% which could introduce decay in leader allocation or leader reallocation during the course of training. 
% This could further enhance the adaptability and performance of LFNNs. 
Additionally, we also investigated the performance of the best-performing leaders in each layer and compared the performance between leaders and followers in the supplementary materials.
% \textbf{Catastrophic forgetting.}

\subsection{BP-free Leader-Follower Neural Networks (LFNN-$\ell$s)}\label{sec:LFNN-l}
In this section, we conduct a comparative analysis between LFNN-$\ell$s and several alternative approaches, with the option of engaging BP. We evaluate their performance on the MNIST, CIFAR-10, and ImageNet datasets to showcase the capabilities of LFNN-$\ell$s and further study the impact of leadership size.
All LFNN-$\ell$s and LFNNs in this section consist of FC and convolutional layers. LFNNs are trained using a combination of BP, global loss, and local losses, while BP-free LFNN-$\ell$s are trained solely with local losses. 

\begin{table}[!t]
\centering
\resizebox{\columnwidth}{!}{%
    \begin{tabular}{cccccc}
    \toprule
    \multicolumn{2}{c|}{\textbf{Dataset}} & \multicolumn{1}{c|}{\textbf{MNIST}}                & \multicolumn{1}{c|}{\textbf{MNIST}}                & \multicolumn{1}{c|}{\textbf{CIFAR-10}}  & \textbf{ImageNet}           \\ 
    % \multicolumn{2}{c|}{Network Scale} & \multicolumn{1}{c|}{S}                    & \multicolumn{1}{c|}{M}                    & \multicolumn{1}{c|} L   & {L}                 \\ 
    \multicolumn{2}{c|}{Metric}  & \multicolumn{1}{c|}{Test / Train Err. ($\downarrow$)} & \multicolumn{1}{c|}{Test / Train Err. ($\downarrow$)} & \multicolumn{1}{c|}{Test / Train Err. ($\downarrow$)} &Test / Train Err. ($\downarrow$) \\ \midrule
     \multirow{3}*{\textbf{BP-enabled}} & \multicolumn{1}{c|}{BP}      & \multicolumn{1}{c|}{2.01 / 0.00}            & \multicolumn{1}{c|}{1.88 / 0.00}            & \multicolumn{1}{c|}{20.90 / 0.00} & 35.24 / 19.14          \\ 
    & \multicolumn{1}{c|}{LG-BP~\cite{belilovsky2019greedy}}   & \multicolumn{1}{c|}{2.43 / 0.00}            & \multicolumn{1}{c|}{2.81 / 0.00}            & \multicolumn{1}{c|}{33.84 / 0.05}  &  54.37 / 39.66      \\ 
    & \multicolumn{1}{c|}{\textbf{LFNN}}    & \multicolumn{1}{c|}{1.18 / 1.15}            & \multicolumn{1}{c|}{2.14 / 1.49}            & \multicolumn{1}{c|}{19.21 / 3.57}   &{57.75 / 20.94}        \\ \midrule 
    % \multicolumn{5}{c}{BP-free}                                                                                                                \\ \hline
     \multirow{6}*{\textbf{BP-free}} & \multicolumn{1}{c|}{FA~\cite{lillicrap2016random}}      & \multicolumn{1}{c|}{2.82 / 0.00}            & \multicolumn{1}{c|}{2.90 / 0.00}            & \multicolumn{1}{c|}{39.94 / 28.44} &  94.55 / 94.13       \\ 
    & \multicolumn{1}{c|}{FG-W~\cite{baydin2022gradients}}    & \multicolumn{1}{c|}{9.25 / 8.93}            & \multicolumn{1}{c|}{8.56 / 8.64}            & \multicolumn{1}{c|}{55.95 / 54.28}    &  97.71 / 97.58   \\ 
    & \multicolumn{1}{c|}{FG-A~\cite{ren2022scaling}}    & \multicolumn{1}{c|}{3.24 / 1.53}            & \multicolumn{1}{c|}{3.76 / 1.75}            & \multicolumn{1}{c|}{59.72 / 41.27}   &  98.83 / 98.80          \\ 
    & \multicolumn{1}{c|}{LG-FG-W~\cite{ren2022scaling}} & \multicolumn{1}{c|}{9.25 / 8.93}            & \multicolumn{1}{c|}{5.66 / 4.59}            & \multicolumn{1}{c|}{52.70 / 51.71}&{97.39 / 97.29}          \\ 
    & \multicolumn{1}{c|}{LG-FG-A~\cite{ren2022scaling}} & \multicolumn{1}{c|}{3.24 / 1.53}            & \multicolumn{1}{c|}{2.55 / 0.00}            & \multicolumn{1}{c|}{30.68 / 19.39}
    & {58.37 / 44.86}\\ 
    & \multicolumn{1}{c|}{\textbf{LFNN-$\ell$}}   & \multicolumn{1}{c|}{\textbf{1.49} / 0.04}            & \multicolumn{1}{c|}{\textbf{1.20} / 1.15}            & \multicolumn{1}{c|}{\textbf{20.85} / 4.69} & \textbf{55.88} / 36.13          \\ \midrule
    \multicolumn{2}{c|}{\textbf{Number of Parameters}} &  \multicolumn{1}{c|}{272K$\sim$275K} & \multicolumn{1}{c|}{429K$\sim$438K} & \multicolumn{1}{c|}{876K$\sim$919K} & \multicolumn{1}{c}{17.3M$\sim$36.8M}\\
    \bottomrule \\
    
    \end{tabular}
}
\caption{Comparison between the proposed model and a set of BP-enabled and BP-free algorithms under MNIST, CIFAR-10, and ImageNet. The best test errors (\%) are highlighted in \textbf{bold}. Leadership size is set to $70\%$ for all the LFNNs and LFNN-$\ell$s.}
\vspace{-.7cm}
\label{tab:res}
\end{table}

\begin{table*}[!t]
    \centering
    % \vspace{-0.3cm}
    \resizebox{\textwidth}{!}{%
    \begin{tabular}{c cc| cccccccccc}\\\toprule  
        %  \specialrule{.15em}{.05em}{.05em} 
         \multirow{1}*{\textbf{Dataset}} &\multicolumn{2}{c|}{\multirow{2}*{\textbf{Model}}} & \multicolumn{10}{c}{\textbf{Leadership Percentage}} \\
          (No. of Parameters)& & & 10\% & 20\% &30\% & 40\% & 50\% & 60\% & 70\% & 80\% &90\% & 100\%    \\
         \midrule

         \multirow{4}*{\textbf{MNIST}} & \multirow{2}*{LFNN-$\ell$} & Test  & {1.96} & 1.67 & 1.98 & 1.49 & 1.49&1.49 & \textbf{1.49} & 1.64 & 1.69 & 1.57\\
         & & Train & {0.12} & 0.42 & 0.07 & 0.06 & 0.05 & 0.11 & 0.04 & 0.36 & 0.24 & 0.92\\
         & \multirow{2}*{LFNN} & Test & 1.24 & 1.68& 1.50 & 1.60 & {1.40} & 1.20 & \textbf{1.18} & 1.40 & 1.44 & 1.51\\
         ($\sim$275K)& & Train & 1.21 & 1.21 & 1.20 & 1.20 & {1.10} & 1.10 & 1.15 & 1.10 & 1.21 & 1.17\\\midrule
         \multirow{4}*{\textbf{MNIST}} & \multirow{2}*{LFNN-$\ell$} & Test & 1.25 & 1.49 & 1.85 & \textbf{1.12} & 1.27 & 1.76 &1.20& 1.64 &1.20&1.23\\
         & & Train & 1.14 & 1.22 & 1.30 & 1.22 & 1.17 & 1.21 & 1.15 & 1.15 & 1.18 & 1.13  \\
         & \multirow{2}*{LFNN} & Test & \textbf{1.89} & 2.49 & 2.20 & 2.97 & 2.23 & 2.70 & 
 2.14 & 2.27 & 2.20 & 2.67 \\
         ($\sim$438K) & & Train & 1.70 & 2.17 & 2.08 & 1.84 & 2.06 & 1.97 & 1.49 & 1.93 & 1.61 & 1.58\\\midrule
         \multirow{4}*{\textbf{CIFAR-10}} & \multirow{2}*{LFNN-$\ell$} & Test & 23.37 & 23.09 & 21.26 & 21.56 & 21.11 & 21.57 & \textbf{20.85} &{21.21}& 21.28 & {21.34}\\
         & & Train & 6.20 & 5.65 & 3.57 & 3.85 & 4.07 & 4.20 & 4.69 & 5.09 & 4.38 & 4.47  \\
         & \multirow{2}*{LFNN} & Test & 23.36 & 20.11 & 19.56 & 19.32 & 19.64 & 18.84 & 
 19.21 & 19.84 & {19.92} & \textbf{18.41} \\
         ($\sim$876K)& & Train & 8.59 & 5.56 & 3.63 & 4.05 & 4.31 & 4.89 & 3.57 & 5.43 & 3.06 & 3.37\\
        \bottomrule
    \end{tabular}
    }
    \caption{Error rate (\% $\downarrow$) results of LFNNs and LFNN-$\ell$s (with different leadership percentage) on MNIST and CIFAR-10.}%
    \vspace{-0.3cm}
    \label{tab:bp-free-cifar}
\end{table*}

\textbf{Datasets.} Both MNIST and CIFAR-10 are obtained from the TensorFlow datasets~\cite{abadi2016tensorflow}. MNIST~\cite{lecun1998mnist} contains 70,000 images, each of size $28\times28$. CIFAR-10~\cite{krizhevsky2009learning} consists of 60,000 images, each of size $32\times32$.
% ImageNet~\cite{deng2009ImageNet} contains 1.3 million images of 1000 classes, which we resized to $224\times224$. 
Tiny ImageNet~\cite{le2015tiny} consists of a dataset of $100,000$ images distributed across 200 classes, with 500 images per class for training, and an additional set of $10,000$ images for testing. All images in the dataset are resized to $64\times64$ pixels. ImageNet subset (1pct)~\cite{chen2020simple, ILSVRC15} is a subset of ImageNet~\cite{deng2009ImageNet}. It shares the same validation set as ImageNet and includes a total of 12,811 images sampled from the ImageNet dataset. These images are resized to $224\times224$ pixels for training.

\textbf{MNIST and CIFAR-10. }We compare our LFNNs and LFNN-$\ell$s with BP, local greedy backdrop (LG-BP)~\cite{belilovsky2019greedy}, Feedback Alignment (FA)~\cite{lillicrap2016random}, weight-perturbed forward gradient (FG-W)~\cite{baydin2022gradients}, activity perturbation forward gradient (FG-A)~\cite{ren2022scaling}, local greedy forward gradient weight / activity-perturbed (LG-FG-W and LG-FG-A)~\cite{ren2022scaling} on MNIST, CIFAR-10, and ImageNet datasets. To ensure a fair comparison, we make slight modifications to our model architectures to match the number of parameters of the models presented in~\cite{ren2022scaling}. 

Table~\ref{tab:res} presents the image classification results for the MNIST and CIFAR-10 datasets using various BP and BP-free algorithms. The table displays the test and train errors as percentages for each dataset and network size.
When comparing to BP-enabled algorithms, LFNN shows similar performance to standard BP algorithms and outperforms the LG-BP algorithm on both the MNIST and CIFAR-10 datasets. In the case of BP-free algorithms, LFNN-$\ell$ achieves lower test errors for both MNIST and CIFAR-10 datasets. Specifically, in MNIST, our LFNN-$\ell$ achieves test error rates of 2.04\% and 1.20\%, whereas the best-performing baseline models achieve 2.82\% and 2.55\%, respectively. For the CIFAR-10 dataset, LFNN-$\ell$ outperforms all other BP-free algorithms with a test error rate of 20.85\%, representing a significant improvement compared to the best-performing LG-FG-A algorithm, which achieves a test error rate of 30.68\%.

In previous sections, we observed that both larger and smaller leadership sizes deliver good performance on simple tasks. This observation holds true for MNIST and CIFAR-10 datasets as shown in Table \ref{tab:bp-free-cifar}. In MNIST, LFNN and LFNN-$\ell$ with different leadership sizes achieve similar test error rates. Further details on the relationship between leadership size and model performance will be discussed in the next subsection.

\textbf{Scaling up to ImageNet.} Traditional BP-free algorithms have shown limited scalability when applied to larger datasets such as ImageNet~\cite{bartunov2018assessing}. To assess the scalability of LFNN and LFNN-$\ell$, we conduct experiments on ImageNet subset and Tiny ImageNet\footnote{More ImageNet results can be found in the supplementary materials.}. The results in Table~\ref{tab:res} compare the train / test error rates of LFNN and LFNN-$\ell$ with other baseline models using BP and BP-free algorithms on the ImageNet dataset.
In the ImageNet experiments, LFNN achieves competitive test errors compared to BP and LG-BP, achieving a test error rate of 57.75\% compared to 35.24\% and 54.37\% respectively. Notably, when compared to BP-free algorithms, LFNN-$\ell$ outperforms all baseline models and achieves a test error rate 2.49\% lower than the best-performing LG-FG-A. Furthermore, LFNN-$\ell$ demonstrates an improvement over LFNN on ImageNet. These results suggest that the use of local loss in LFNN-$\ell$ yields better performance compared to global loss, particularly when dealing with challenging tasks such as ImageNet.

To further investigate the generalizability of LFNN and LFNN-$\ell$, we conduct experiments on ImageNet variants and increase the model size by doubling the number of parameters to approximately 37M. Additionally, we explore the impact of leadership size on model performance. The results of the error rates for Tiny ImageNet and ImageNet subset with varying leadership percentages are presented in Table~\ref{tab:bp-free_tiny}.
For Tiny ImageNet, we observe that using a leadership percentage of 90\% yields the lowest test error rates, with LFNN achieving 35.21\% and LFNN-$\ell$ achieving 36.06\%. These results are surprisingly comparable to other BP-enabled deep learning models tested on Tiny ImageNet,  such as UPANets (test error rate $=32.33\%$)~\cite{tseng2022upanets}, PreActRest (test error rate $=36.52\%$)~\cite{kim2020puzzle}, DLME (test error rate $=55.10\%$)~\cite{zang2022dlme}, and MMA (test error rate $=35.59\%$)~\cite{konstantinidis2022multi}.

In the ImageNet subset experiments, we follow the methodology of \cite{chen2020simple} and leverage the ResNet-50 architecture as the base encoder, combining it with LFNN and LFNN-$\ell$.
LFNN and LFNN-$\ell$ with 90\% leadership achieve the lowest test error rates of 57.37\% and 53.82\%, respectively. These results surpass all baseline models in Table \ref{tab:res} and are even comparable to the test error rates of BP-enabled algorithms reported in \cite{chen2020simple}, which is 50.6\%. This observation further demonstrates the effectiveness of our proposed algorithm in transfer learning scenarios.
% Notably, the best test error rate under the ImageNet subset in Table \ref{tab:bp-free_tiny} is 53.82\%, achieved with 90\% leadership in a network with approximately 37M parameters. 
It is worth mentioning that we observed even better results than those in Table \ref{tab:bp-free_tiny} when further increasing the number of parameters. From Figure \ref{fig:pmnist}a and Table \ref{tab:bp-free-cifar}, we recall that for simple tasks like MNIST or CIFAR-10 classification, small leadership sizes can achieve decent results. In Table \ref{tab:bp-free_tiny}, we observe a clearer trend that for difficult datasets like ImageNet, a higher leadership percentage is required to achieve better results. This presents an interesting avenue for future exploration, particularly in understanding the relationship between network / leadership size and dataset complexity.

% \begin{table*}[!t]
%     \centering
%     % \vspace{-0.3cm}
%     \resizebox{\textwidth}{!}{%
%     \begin{tabular}{c c cccccccccc}\\\toprule  
%         %  \specialrule{.15em}{.05em}{.05em} 
%                  Dataset & Model & 10\% & 20\% &30\% & 40\% & 50\% & 60\% & 70\% & 80\% &90\% & 100\%    \\
%                \midrule

%          \multirow{4}*{ImageNet-T} & LFNN-$\ell$ (Test)  & {73.98\%} & 63.09\% & 54.24\% & 49.63\% & 44.87\% & 40.96\% & 37.17\% & 38.05\% & \textbf{36.06\%} & 39.56\\
%          & LFNN-$\ell$ (Train) & {71.47\%} & 57.29\% & 43.69\% & 38.57\% & 30.53\% & 22.04\% & 19.50\% & 19.38\% & 16.00\% & 32.33\\
%          & LFNN (Test) & 39.85\% & 40.12\% & 39.34\% & 39.18\% & {39.33\%} & 39.41\% & 39.42\% & 38.63\% & \textbf{35.21\%} & 39.56\\
%          & LFNN (Train) & 36.50\% & 35.76\% & 32.71\% & 32.16\% & {32.02\%} & 32.36\% & 32.70\% & 31.91\% & 32.59\% & 32.33\\\hline
%          \multirow{4}*{ImageNet-S} & LFNN-$\ell$ (Test) & \\
%          & LFNN-$\ell$ (Train) & \\
%          & LFNN (Test) & \\
%          & LFNN (Train) &\\
%         \bottomrule
%     \end{tabular}
%     }
%     \caption{\textcolor{black}{Error rate results of LFNNs (with different leadership percentage) trained with and without BP (LFNN-$\ell$) on Tiny ImageNet (ImageNet-T) and ImageNet subset (ImageNet-S). The CNN BP test error for ImageNet-T and ImageNet-S are 30.46\% and 35.24\%, respectively. }
%     }%
%     % \vspace{-0.5cm}
%     \label{tab:bp-free_tiny}
% \end{table*}

\begin{table*}[!t]
    \centering
    % \vspace{-0.3cm}
    \resizebox{\textwidth}{!}{%
    \begin{tabular}{c cc| cccccccccc}\\\toprule  
        %  \specialrule{.15em}{.05em}{.05em} 
         \multirow{2}*{\textbf{Dataset}} &\multicolumn{2}{c|}{\multirow{2}*{\textbf{Model}}} & \multicolumn{10}{c}{\textbf{Leadership Percentage}} \\
          & & & 10\% & 20\% &30\% & 40\% & 50\% & 60\% & 70\% & 80\% &90\% & 100\%    \\
         \midrule

         \multirow{4}*{\textbf{Tiny ImageNet}} & \multirow{2}*{LFNN-$\ell$} & Test  & {73.98} & 63.09 & 54.24 & 49.63 & 44.87 & 40.96 & 37.17 & 38.05 & \textbf{36.06} & 39.56\\
         & & Train & {71.47} & 57.29 & 43.69 & 38.57 & 30.53 & 22.04 & 19.50 & 19.38 & 16.00 & 32.33\\
         & \multirow{2}*{LFNN} & Test & 39.85 & 40.12& 39.34 & 39.18 & {39.33} & 39.41 & 39.42 & 38.63 & \textbf{35.21} & 39.56\\
         & & Train & 36.50 & 35.76 & 32.71 & 32.16 & {32.02} & 32.36 & 32.70 & 31.91 & 32.59 & 32.33\\\midrule
         \multirow{4}*{\textbf{ImageNet Subset}} & \multirow{2}*{LFNN-$\ell$} & Test & 90.57 & 84.83 & 78.75 & 73.65 & 68.61 & 64.25 &59.53& 56.54 &\textbf{53.82}&54.44\\
         & & Train & 68.96 & 51.89 & 39.49 & 27.78 & 22.68 & 13.37 & 9.23 & 5.41 & 5.58 & 6.40  \\
         & \multirow{2}*{LFNN} & Test & 79.37 & 78.83 & 69.87 & 61.80 & 60.05 & 59.10 & 
 57.46 &58.01 & \textbf{57.37} &57.75 \\
         & & Train & 53.13 & 52.18 & 38.38 & 26.26 & 25.21 & 20.35 & 18.42 & 18.40 & 16.70 & 17.94\\
        \bottomrule
    \end{tabular}
    }
    \caption{\textcolor{black}{Error rate (\% $\downarrow$) results of LFNNs and LFNN-$\ell$s (with different leadership percentage) on Tiny ImageNet and ImageNet subset. We also trained CNN counterparts (without LF hierarchy) with BP and global loss for reference. The test error rates of BP-enabled CNNs under Tiny ImageNet and ImageNet subset are 35.76\% and 51.62\%, respectively. }
    }%
    \vspace{-0.5cm}
    \label{tab:bp-free_tiny}
\end{table*}

% \textcolor{red}{This architecture allows you to mimic the properties of neuronal networks where it is energy inefficient and computationally expensive to have global transmission of the error signal, and also unnecessary. Maybe you can say that. Since each neuron has a more complex structure (leader + followers), it seems that each one is like a mini-NN. So you are essentially encapsulating these mini-NNs into a larger NN, and maybe that can help with explainability and interpretability of what each part of the larger NN is doing. So maybe a natural extension of this is to adversarial networks, or at least to networks that can solve adversarial problems locally rather than by inefficiently talking to each other back and forth. I don't know if this makes sense, but there seems to be potential in the scalability of the leader-follower concept.}

% \textcolor{blue}{mingxi: yes overall you are right about the several optimization. I've done tons of experiments on that, and adversarial attack defence, continual learning, OOD generalization, tried to solve catastrophic forgetting, but unfortunately results are no better than normal BP :( Probably we can put one sentence discussion in conclusion and explore more in our future works.}

% \textcolor{red}{I suggest asking ChatGPT to fix punctuation and grammar errors throughout the text. These errors are distracting and annoy reviewers.}
\vspace{-3mm}
\section{Conclusion}
\vspace{-2mm}
In this work, we have presented a novel learning algorithm, LFNN, inspired by collective behavior observed in nature. By introducing a leader-follower hierarchy within neural networks, we have demonstrated its effectiveness across various network architectures. Our comprehensive study of LFNN aligns with observations and theoretical foundations in both the biological and deep learning domains.
In addition, we have proposed LFNN-$\ell$, a BP-free variant that utilizes local error signals instead of traditional backpropagation. We have shown that LFNN-$\ell$, trained without a global loss, achieves superior performance compared to a set of BP-free algorithms. Through extensive experiments on MNIST, CIFAR-10, and ImageNet datasets, we have validated the efficacy of LFNN with and without BP. LFNN-$\ell$ not only outperforms other state-of-the-art BP-free algorithms on all tested datasets but also achieves competitive results when compared to BP-enabled baselines in certain cases.
Our work is unique as it is the first to introduce collective motion-inspired models for deep learning architectures, opening up new directions for the development of local error signals and alternatives to BP. The proposed algorithm is straightforward yet highly effective, holding potential for practical applications across various domains. We believe that this early study provides valuable insights into fundamental challenges in deep learning, including neural network architecture design and the development of biologically plausible decentralized learning algorithms.

\clearpage

\bibliography{neurips_2023}
\bibliographystyle{unsrt}

\end{document}